\documentclass{article}
\usepackage{spconf,amsmath,graphicx,multirow}
\usepackage{times}
\usepackage{comment}
\usepackage{color}
\includecomment{comment}

\title{Achieving Human Parity in Conversational Speech Recognition}
\name{W. Xiong, J. Droppo, X. Huang, F. Seide, M. Seltzer, A. Stolcke, D. Yu and G. Zweig}
\address{Microsoft Research\\
Technical Report MSR-TR-2016-71\\
Revised February 2017}
\begin{document}
\maketitle
\begin{abstract}
Conversational speech recognition has 
served as a flagship speech recognition task since the release of the
Switchboard corpus in the 1990s. In this paper, we measure the
human error rate on the widely used NIST 2000 test set, and find that
our latest automated system has reached human parity. The error rate
of professional transcribers is 5.9\% for the Switchboard portion of the data, in which
newly acquainted pairs of people discuss an assigned topic, and 11.3\% for
the CallHome portion where friends and family members have open-ended
conversations. In both cases, our automated system establishes a 
new state of the art, and edges past the human
benchmark, achieving error rates of 5.8\% and 11.0\%, respectively.
The key to our system's performance is the
use of various convolutional and LSTM acoustic model architectures, combined with a novel
spatial smoothing method and lattice-free MMI acoustic training,
multiple recurrent neural network language modeling approaches, 
and a systematic use of system combination.
\end{abstract}
\section{Introduction}
\label{sec:intro}
Recent years have seen human performance levels reached or surpassed in 
tasks ranging from the games of chess and Go 
\cite{campbell2002deep,silver2016mastering} 
to simple speech recognition 
tasks like carefully read newspaper 
speech \cite{amodei2015deep} and rigidly constrained small-vocabulary tasks in noise \cite{kristjansson2006super,weng2014single}. In the area of 
speech recognition, much of the
pioneering early work was driven by a series of carefully designed 
tasks with DARPA-funded 
datasets publicly released 
by the LDC and NIST \cite{pallett2003look}: first simple ones like 
the ``resource management'' task \cite{price1988darpa} with a small vocabulary and carefully
controlled grammar; then read speech recognition in the Wall Street 
Journal task \cite{paul1992design}; then Broadcast News \cite{graff19971996}; each progressively
more difficult for automatic systems. One of last big initiatives in this area was
in conversational telephone speech (CTS), which is especially difficult due to the
spontaneous (neither read nor planned) nature of the speech, its informality,
and the self-corrections, hesitations and other disfluencies that are pervasive.
The Switchboard \cite{godfrey1992switchboard} and later Fisher \cite{cieri2004fisher} data collections of the 1990s and early 2000s provide what is to date 
the largest and best studied of the conversational corpora.
The history of work in this area includes key contributions by institutions
such as IBM \cite{chen2006advances}, BBN \cite{matsoukas2006advances},
SRI \cite{stolcke2006recent}, AT\&T \cite{ljolje2001}, LIMSI \cite{gauvain2003conversational}, Cambridge University \cite{evermann2004development},
Microsoft \cite{seide2011conversational}
and numerous others. 
In the past, human performance on this task has been widely cited as
being 4\% \cite{lippmann1997speech}. However, the error rate estimate 
in \cite{lippmann1997speech} is
attributed to a ``personal communication,'' and the actual source of this
number is ephemeral. To better understand human performance, 
we have used professional transcribers to transcribe 
the actual test sets that we are working with,
specifically the CallHome and Switchboard portions of the NIST eval 2000
test set. We find that the human error rates on these two parts are 
different almost by a factor of two, so a single number is 
inappropriate to cite. The error rate on Switchboard is about  5.9\%,
and for CallHome 11.3\%. We improve on our recently reported
conversational speech recognition system \cite{ms-swb-icassp2017} by about 0.4\%, and now 
exceed human performance by a small margin. 
Our progress is a result of the careful engineering and optimization
of convolutional and recurrent neural networks. While the 
basic structures have been well known
for a long period \cite{pineda1987generalization,williams1989learning,waibel1989phoneme,lecun1995convolutional,lecun1989backpropagation,robinson1991recurrent,hochreiter1997long},
it is only recently that they have dominated the field as the best models for
speech recognition. Surprisingly, this is the case for both acoustic
modeling \cite{sak2014long,sak2015fast,saon2015ibm,sercu2016very,bi2015very,qian2016very} 
and language modeling \cite{mikolov2010recurrent,mikolov2012context,sundermeyer2012lstm,medennikov2016improving}. 
In comparison to the standard feed-forward MLPs or DNNs
that first demonstrated breakthrough performance on conversational 
speech recognition \cite{seide2011conversational}, these acoustic models
have the ability to model a large amount of acoustic context with temporal
invariance, and in the case of convolutional models, with frequency invariance
as well.
In language modeling, recurrent models appear to improve over classical N-gram models
through the use of an unbounded word history, as well as 
the generalization ability of continuous word representations \cite{mikolov2013linguistic}.
In the meantime, ensemble learning has 
become commonly used in several neural models \cite{sutskever2014sequence,hannun2014deep,mikolov2012context},
to improve robustness by reducing bias and variance. 
This paper is an expanded version of \cite{ms-swb-icassp2017}, with the following
additional material:
\begin{enumerate}
\item A comprehensive assessment of human performance on the 
NIST eval 2000 test set
\item The description of a novel spatial regularization method which significantly boosts our LSTM acoustic model
performance
\item The use of LSTM rather than RNN-LMs, and the use of a letter-trigram 
input representation
\item A two-level system combination, based on a subsystem of BLSTM-system variants that, by itself, surpasses the 
best previously reported results
\item Expanded coverage of the Microsoft Cognitive Toolkit (CNTK) used to
build our models
\item An analysis of human versus machine errors, which indicates substantial equivalence, with the exception of the word classes of
backchannel acknowledgments (e.g. ``uh-huh'') and hesitations  (e.g. ``um''). 
\end{enumerate}
The remainder of this paper describes our system in detail.
Section \ref{sec:human} describes our measurement of human performance. 
Section \ref{sec:cnn+lstm} describes the convolutional neural net (CNN) and long-short-term memory (LSTM) models.
Section \ref{sec:sam} describes our implementation of i-vector adaptation. Section \ref{sec:lfmmi}
presents out lattice-free MMI training process.
Language model rescoring is a significant part of our system, and described in Section \ref{sec:rescoring}.
We describe the CNTK toolkit that forms the basis of our neural network models in Section \ref{sec:cntk}. 
Experimental results are presented in Section \ref{sec:results}, followed
by an error analysis in section \ref{sec:analysis}, 
a review of relevant prior work in \ref{sec:prior} and concluding remarks.
\section{Human Performance}
\label{sec:human}
To measure human performance, we leveraged an existing pipeline in which 
Microsoft data is transcribed on a weekly basis. This pipeline uses 
a large commercial vendor to perform two-pass transcription. In the first pass,
a transcriber works from scratch to transcribe the data. In the second pass,
a second listener monitors the data to do error correction. Dozens of hours
of test data are processed in each batch. 
One week, we added the NIST 2000 CTS evaluation data to the work-list, without 
further comment. The intention was to measure the error rate of 
professional transcribers going about their normal everyday business.
Aside from the standard two-pass checking in place, we did not do 
a complex multi-party transcription and adjudication 
process. The transcribers were given the same audio segments as were
provided to the speech recognition system, which results in 
short sentences or sentence fragments from a single channel. This makes
the task easier since the speakers are more clearly separated, and more
difficult since the two sides of the conversation are not interleaved. Thus, it 
is the same condition as reported for our automated systems. 
The resulting numbers are 5.9\% for the Switchboard portion,
and 11.3\% for the CallHome portion of the NIST 2000 test set,
using the NIST scoring protocol.
These numbers should be
taken as an indication of the ``error rate'' of a trained professional 
working in industry-standard speech transcript production.
(We have submitted the human transcripts thus produced to the Linguistic Data Consortium
for publication, so as to facilitate research by other groups.)
Past work \cite{glenn2010transcription} reports inter-transcriber error
rates for data taken from the later
RT03 test set (which contains Switchboard and Fisher, but no CallHome data).
Error rates of 4.1 to 4.5\% are reported for extremely careful multiple
transcriptions, and 9.6\% for ``quick transcriptions.''
While this is a different test set, the numbers are in line with 
our findings. 
We note that the bulk of the Fisher training data, and 
the bulk of the data overall, was transcribed with the ``quick transcription''
guidelines. Thus, the current state of the art is actually far exceeding the 
noise level in its own training data.
Perhaps the most important point is the extreme  variability between the two test subsets.
The more informal CallHome data has almost double the human error 
rate of the Switchboard data. Interestingly, the same informality, multiple speakers per channel,
and recording conditions that make CallHome hard for computers
make it difficult for people as well.  Notably, the performance of 
our artificial system aligns almost exactly with the performance of people
on both sets.
\section{Convolutional and LSTM Neural Networks}
\label{sec:cnn+lstm}
\subsection{CNNs}
\begin{figure}[t]
\centering
\includegraphics[width=0.45\textwidth]{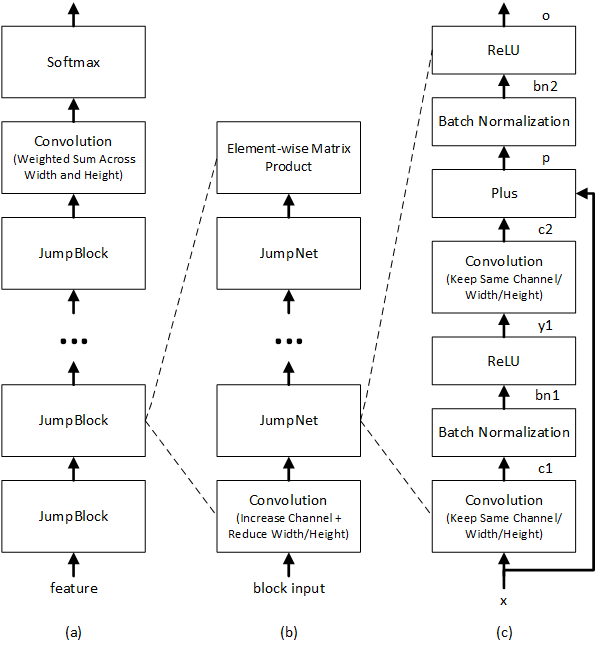}
\caption{LACE network architecture}
\label{fig:LACE}
\end{figure}
We use three CNN variants. The first is the VGG architecture of \cite{Simonyan2014very}. Compared to the networks used previously in image recognition, this 
network uses small (3x3) filters, is deeper, and applies up to 
five convolutional layers before pooling. 
The second network is modeled on the ResNet architecture \cite{he2015deep}, 
which adds highway connections \cite{DBLP:journals/corr/SrivastavaGS15}, i.e. a linear transform of each layer's input to the layer's output
\cite{DBLP:journals/corr/SrivastavaGS15,ghahremani2016linearly}. The only difference is that we apply Batch Normalization before computing ReLU activations.
The last CNN variant is the LACE (layer-wise context expansion with attention) model \cite{yu2016deep}. LACE is a TDNN \cite{waibel1989phoneme} variant in which each higher layer is a weighted sum of nonlinear transformations of a window of lower layer frames. In other words, each higher layer exploits broader context than lower layers. Lower layers focus on extracting simple local patterns while higher layers extract complex patterns that cover broader contexts.
Since not all frames in a window carry the same importance, an attention mask is applied.
The LACE model differs from the earlier TDNN models e.g. \cite{waibel1989phoneme,waibel1989consonant} in the use of a learned attention mask and ResNet like
linear pass-through. As illustrated in detail in Figure \ref{fig:LACE}, the model is composed of four blocks, each with the same architecture. Each block starts with a convolution layer with stride 2 which sub-samples the input and increases the number of channels. This layer is followed by four RELU-convolution layers with jump links similar to those used in ResNet. 
Table \ref{tab:CNNs} compares the layer structure and parameters of the three CNN architectures. 
We also trained a fused model by combining a ResNet model and a VGG model at 
the senone posterior level. Both base models are independently trained, and then the score fusion weight is optimized on development data. 
The fused system is our best single system. 
\begin{table}[t]
\centering
\caption{Comparison of CNN architectures}
\vspace*{0.1in}
\label{tab:CNNs}
\includegraphics[width=0.45\textwidth]{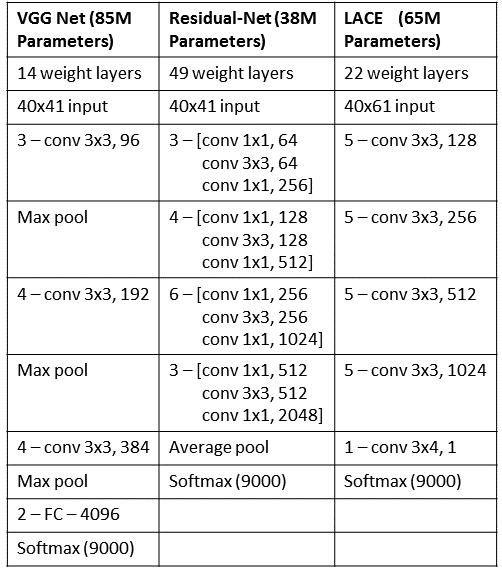}
\end{table}
\subsection{LSTMs}
While our best performing models are convolutional, the
use of long short-term memory networks (LSTMs) is a close second. We use a 
bidirectional architecture \cite{graves2005framewise} without frame-skipping
\cite{sak2015fast}. The core model structure is the LSTM defined in 
\cite{sak2014long}. We found that using networks with more than six
layers did not improve the word error rate on the development set,
and chose 512 hidden units, per direction, per layer, as that provided a 
reasonable trade-off between training time and final model accuracy.
\subsection{Spatial Smoothing}
Inspired by the human auditory cortex, where neighboring neurons tend to simultaneously activate, we employ a spatial smoothing technique to improve the accuracy of our LSTM models.
The smoothing is implemented as a regularization term on the activations between layers of the acoustic model.
First, each vector of activations is re-interpreted as a 2-dimensional image. For example, if there are 512 neurons, they are interpreted as the pixels of a 16 by 32 image.
Second, this image is high-pass filtered. The filter is implemented as a circular convolution with a 3 by 3 kernel. The center tap of the kernel has a value of $1$, and the remaining eight having a value of $-1/8$.
Third, the energy of this high-pass filtered image is computed and added to the training objective function. We have found empirically that a suitable scale for this energy is $0.1$ with respect to the existing cross entropy objective function.
The overall effect of this process is to make the training algorithm prefer models that have correlated neurons, and to improve the word error rate of the acoustic model.
Table~\ref{tab:spatialsmoothing} shows the benefit in error rate for some of our early systems.
We observed error reductions of between 5 and 10\% relative from spatial smoothing.
\begin{table}[t]
    \centering
\caption{Accuracy improvements from spatial smoothing on the NIST 2000 CTS test set.
The model is a six layer BLSTM, using i-vectors and 40 dimensional filterbank features, and a dimension of 512 in each direction of each layer.}
\label{tab:spatialsmoothing}
\vspace*{0.1in}
    \begin{tabular}{|l|c|c|c|c|}
    \hline 
	\multirow{3}{*}{Model} & \multicolumn{4}{c|}{WER (\%)}	\\ 
				\cline{2-5}
				& \multicolumn{2}{c|}{9000 senones} & \multicolumn{2}{c|}{27000 senones} \\
				\cline{2-5}
				& CH	& SWB	& CH	& SWB	\\ \hline
	Baseline       		& 21.4	& 9.9	& 20.5	& 10.6	\\ \hline
	Spatial smoothing    	& 19.2	& 9.3	& 19.5	& 9.2	\\ \hline
    	\end{tabular}
\end{table}
\section{Speaker Adaptive Modeling}
\label{sec:sam}
Speaker adaptive modeling in our system is based on 
conditioning the network on an i-vector  \cite{dehak2011front}  
characterization of each 
speaker \cite{saon2013speaker,saonSRK16}.
A 100-dimensional i-vector is generated for each conversation side.
For the LSTM system,
the conversation-side i-vector $v_s$ is appended to each frame of input.
For convolutional networks, this approach is inappropriate because
we do not expect to see spatially contiguous patterns in the input.
Instead, for the CNNs, we add a learnable weight matrix $W^l$ to each 
layer, and add $W^l v_s$ to the activation of the layer before the 
nonlinearity. Thus, in the CNN, the i-vector essentially serves as an 
speaker-dependent bias to each layer. Note that the i-vectors are estimated
using MFCC features; by using them subsequently in systems based on
log-filterbank features, we may benefit from a form of feature 
combination.
Performance improvements from i-vectors are shown in Table \ref{tab:lfmmi}. 
The full experimental setup is described in Section \ref{sec:results}.
\begin{table*}[t]
    \centering
\caption{Performance improvements from i-vector and LFMMI training on the NIST 2000 CTS test set}
\vspace{0.1in}
\label{tab:lfmmi}
        \small
    \begin{tabular}{|l|c|c|c|c|c|c|c|c|}
    \hline
        \multirow{3}{*}{Configuration} & \multicolumn{8}{c|}{WER (\%)}                                  \\ \cline{2-9}
                               & \multicolumn{2}{c|}{ReLU-DNN} & \multicolumn{2}{c|}{ResNet-CNN}
				& \multicolumn{2}{c|}{BLSTM} & \multicolumn{2}{c|}{LACE}\\ \cline{2-9}
                                & CH    & SWB   & CH	& SWB	& CH    & SWB   & CH    & SWB   \\ \hline
        Baseline                & 21.9  & 13.4  & 17.5	& 11.1	& 17.3  &  10.3 & 16.9  & 10.4          \\ \hline
        i-vector                & 20.1  & 11.5  & 16.6	& 10.0	& 17.6  &  9.9  & 16.4  & 9.3     \\ \hline
        i-vector+LFMMI          & 17.9  & 10.2  & 15.2	& 8.6	& 16.3  &  8.9  & 15.2  & 8.5           \\ \hline
        \end{tabular}
\end{table*}
\section{Lattice-Free Sequence Training}
\label{sec:lfmmi}
After standard cross-entropy training, we optimize the model parameters using
the maximum mutual information (MMI) objective function. Denoting a word sequence by $w$ and its 
corresponding acoustic realization by $a$, the training criterion is
\[
	\sum_{w,a \in \text{data}} \log \frac{P(w) P(a|w)} {\sum_w' P(w') P(a|w')}	\quad .
\]
As noted in \cite{sim2010sequential,vesely2013sequence},
the necessary gradient for use in backpropagation is a simple function 
of the posterior probability of a  particular acoustic model state at a given
time, as computed by summing over all possible word sequences in an 
unconstrained manner.  As first done in \cite{chen2006advances},  and more
recently in \cite{povey2016purely}, this can be accomplished with a 
straightforward alpha-beta computation over the finite state acceptor 
representing the decoding search space. In \cite{chen2006advances}, the 
search space is taken to be an acceptor representing the composition $HCLG$ for a 
unigram language model $L$ on words.
In \cite{povey2016purely}, a language model on phonemes is used instead.
In our implementation, we use a mixed-history acoustic unit language model.
In this model, the probability of transitioning into a new context-dependent phonetic state (senone)
is conditioned on both the senone and phone history.
We found this model to perform better than either purely word-based or phone-based models.
Based on a set of initial experiments, we developed the following procedure:
\begin{enumerate}
\item Perform a forced alignment of the training data to select 
lexical variants and determine frame-aligned senone sequences.
\item Compress consecutive framewise occurrences of a single senone into a single occurrence. 
\item Estimate an unsmoothed, variable-length N-gram language model from this data, where the history state consists of the previous phone and previous senones within the current phone.
\end{enumerate}
\newcommand{\phone}[1]{{\it #1}}
\newcommand{\state}[2]{{\it #1}\_s#2}
To illustrate this, consider the sample senone sequence
\{\state{s}{2.1288}, \state{s}{3.1061}, \state{s}{4.1096}\},
\{\state{eh}{2.527}, \state{eh}{3.128}, \state{eh}{4.66}\},
\{\state{t}{2.729}, \state{t}{3.572}, \state{t}{4.748}\}.
When predicting the state following \state{eh}{4.66} the history consists of
(\phone{s}, \state{eh}{2.527}, \state{eh}{3.128}, \state{eh}{4.66}),
and following \state{t}{2.729}, the history is (\phone{eh}, \state{t}{2.729}).
We construct the denominator graph from this language model, and HMM transition
probabilities as determined by transition-counting in the 
senone sequences found in the training data. Our approach not only largely reduces 
the complexity of building up the language model but also provides 
very reliable training performance. 
We have found it convenient to do the full computation, without pruning, in a series of matrix-vector
operations on the GPU. The underlying acceptor is represented with a 
sparse matrix, and we maintain a dense likelihood vector for each time
frame. The alpha and beta recursions are implemented with CUSPARSE level-2
routines: sparse-matrix, dense vector multiplies. Run time is about 100 times
faster than real time.
As in \cite{povey2016purely}, we use 
cross-entropy regularization. 
In all the lattice-free MMI (LFMMI) experiments mentioned below we use a trigram language model.
Most of the gain is usually obtained after processing 24 to 48 hours of data.
\section{LM Rescoring and System Combination}
\label{sec:rescoring}
An initial decoding is done with a WFST decoder, 
using the architecture described in \cite{mendis2016parallelizing}.
We use an N-gram language model trained and pruned with the SRILM toolkit \cite{stolcke2002srilm}.
The first-pass LM has approximately 15.9 million bigrams, trigrams, and 4grams, and a vocabulary of 30,500 words.
It gives a perplexity of 69 on the 1997 CTS evaluation transcripts.
The initial decoding produces a lattice with the pronunciation variants
marked, from which 500-best lists are generated for rescoring purposes.
Subsequent N-best rescoring uses an unpruned LM comprising 145 million N-grams.
All N-gram LMs were estimated by a maximum entropy criterion as
described in \cite{AlumaeKurimo:interspeech2012}.
The N-best hypotheses are then rescored using a combination of the large N-gram LM and several 
neural net LMs. We have experimented with both RNN LMs and LSTM LMs, 
and describe the details in the following two sections.
\subsection{RNN-LM setup}
Our RNN-LMs are
trained and evaluated using the CUED-RNNLM toolkit \cite{chen2016cued}.
Our RNN-LM configuration has several distinctive features, as described below.
\begin{enumerate}
\item
	We trained both standard, forward-predicting RNN-LMs and backward RNN-LMs that predict words
	in reverse temporal order.
	The log probabilities from both models are added.
\item
	As is customary, the RNN-LM probability estimates are interpolated at the word-level with
	corresponding N-gram LM probabilities (separately for the forward and backward models).
	In addition, we trained a second RNN-LM for each direction, obtained by starting with different random 
	initial weights.
	The two RNN-LMs and the N-gram LM for each direction are interpolated with weights of (0.375, 0.375, 0.25).
\item
	In order to make use of LM training data that is not fully matched to the target conversational speech domain,
	we start RNN-LM training with the union of in-domain (here, CTS) and out-of-domain (e.g., Web) data.
	Upon convergence, the
	network undergoes a second training phase using the in-domain data only. Both training phases use in-domain
	validation data to regulate the learning rate schedule and termination.
	Because the size of the out-of-domain data is a multiple of the in-domain data, a standard training
	on a simple union of the data
	would not yield a well-matched model, and have poor perplexity in the target domain.
\item
	We found best results with an RNN-LM configuration that had a second, non-recurrent hidden layer.
	This produced lower perplexity and word error than the standard, single-hidden-layer RNN-LM architecture
	\cite{mikolov2010recurrent}.\footnote{However, adding more hidden layers produced no further gains.}
	The overall network architecture thus had two hidden layers with 1000 units each, using ReLU nonlinearities.
	Training used noise-contrastive estimation (NCE) \cite{Gutmann:NCE}.
\item
	The RNN-LM output vocabulary consists of all words occurring more than once in the in-domain training set.
	While the RNN-LM estimates a probability for unknown words, we take a different approach in rescoring:
	The number of out-of-set words is recorded for each hypothesis and a penalty for them is estimated for them
	when optimizing the relative weights for all model scores (acoustic, LM, pronunciation),
	using the SRILM {\em nbest-optimize} tool.
\end{enumerate}
\subsection{LSTM-LM setup}
After obtaining good results with RNN-LMs we also explored the LSTM recurrent network architecture for
language modeling, inspired by recent work showing gains over RNN-LMs for conversational speech recognition
\cite{medennikov2016improving}.
In addition to applying the lessons learned from our RNN-LM experiments, we explored additional alternatives,
as described below.
\begin{enumerate}
\item
	There are two types of input vectors our LSTM LMs take, word based one-hot vector input and letter trigram vector \cite{huang2013learning} input.
	Including both forward and backward models, we trained four different LSTM LMs in total. 
\item
	For the word based input, we leveraged the approach from \cite{press2016using} to tie the input embedding and output embedding together. 
\item
	Here we also used a two-phase training schedule to train the LSTM LMs. First we train the model on the combination of in-domain 
	and out-domain data for four data passes without any learning rate adjustment. We then start from the resulting model 
	and train on in-domain data until convergence. 
\item
	Overall the letter trigram based models perform a little better than the word based language model. We tried applying dropout on both types of language
	models but didn't see an improvement.
\item
	Convergence was improved through a variation of self-stabilization \cite{ghahremani2016stab}, in which each output vector $x$ of non-linearities are scaled by
	$\frac{1}{4}\ln(1 + e^{4\beta})$, where a $\beta$ is a scalar that is learned for each output.
	This has a similar effect as the scale of the well-known batch normalization technique, but can be used in recurrent loops.
\item
	Table \ref{tab:lstmlm-layers} shows the impact of number of layers on the final perplexities.
	Based on this, we proceeded with three hidden layers, with 1000 hidden units each.
	The perplexities of each LSTM-LM we used in the final combination (before interpolating with the N-gram model)
	can be found in Table \ref{tab:lstmlm-perfs}.
\end{enumerate}
\begin{table}[t]
    \centering
    \caption{LSTM perplexities (PPL) as a function of hidden layers, trained on in-domain data only, computed on 1997 CTS eval transcripts.}
\vspace*{0.1in}
	\label{tab:lstmlm-layers}
    \begin{tabular}{|l|c|}
    \hline
	Language model					&  PPL \\
     \hline \hline
	letter trigram input with one layer (baseline) & 63.2 \\ \hline
	\quad + two hidden layers & 61.8 \\ \hline
	\quad + three hidden layers & 59.1 \\ \hline
	\quad + four hidden layers & 59.6 \\ \hline
	\quad + five hidden layers & 60.2 \\ \hline
	\quad + six hidden layers & 63.7 \\ \hline
    \end{tabular}
\vspace*{-0.1in}
\end{table}
\begin{table}[t]
    \centering
    \caption{Perplexities (PPL) of the four LSTM LMs used in the final combination. PPL is computed on 1997 CTS eval transcripts. All the LSTM LMs are with three hidden layers.}
\vspace*{0.1in}
	\label{tab:lstmlm-perfs}
    \begin{tabular}{|l|c|}
    \hline 
	Language model					&  PPL \\
     \hline \hline
	RNN: 2 layers + word input  (baseline) & 59.8 \\ \hline
	LSTM: word input in forward direction & 54.4 \\ \hline
	LSTM: word input in backward direction & 53.4 \\ \hline
	LSTM: letter trigram input in forward direction & 52.1 \\ \hline
	LSTM: letter trigram input in backward direction & 52.0 \\ \hline
    \end{tabular}
\vspace*{-0.1in}
\end{table}
\begin{table}
    \centering
    \caption{Performance of various versions of neural-net-based LM rescoring.
		Perplexities (PPL) are computed on 1997 CTS eval transcripts;
		word error rates (WER) on the NIST 2000 Switchboard test set.}
\vspace*{0.1in}
	\label{tab:rnnlm-results}
    \begin{tabular}{|l|c|c|}
    \hline
	Language model					&  PPL	& WER \\
     \hline \hline
       4-gram LM (baseline)                            & 69.4  & 8.6 \\
      \hline
       \  + RNNLM, CTS data only                       & 62.6  & 7.6 \\
	\quad + Web data training                      & 60.9  & 7.4 \\
	\quad \quad + 2nd hidden layer                 & 59.0  & 7.4 \\
	\quad \quad \quad + 2-RNNLM interpolation      & 57.2  & 7.3 \\
	\quad \quad \quad \quad + backward RNNLMs      & -     & 6.9 \\
      \hline
	+ LSTM-LM, CTS + Web data		& 51.4	& 6.9 \\	% expt216b
	\quad + 2-LSTM-LM interpolation		& 50.5	& 6.8 \\	% expt217
	\quad \quad + backward LSTM-LM		& -	& 6.6 \\	% expt218
      \hline
    \end{tabular}
\end{table}
For the final system, we interpolated two LSTM-LMs with an N-gram LM for the forward-direction LM,
and similarly for the backward-direction LM.
All LSTMs use three hidden layers and are trained on in-domain and web data. 
Unlike for the RNN-LMs, the two models being interpolated differ not just in their random initialization, but 
also in the input encoding (one uses a triletter encoding, the other a one-hot word encoding).
The forward and backward LM log probability scores are combined additively.
\subsection{Training data}
The 4-gram language model for decoding was trained on the available CTS transcripts from the 
DARPA EARS program: Switchboard (3M words), BBN Switchboard-2 transcripts (850k), Fisher (21M), 
English CallHome (200k), and the University of Washington conversational Web corpus (191M).
A separate N-gram model was trained from each source and interpolated with weights optimized on RT-03 transcripts.
For the unpruned large rescoring 4-gram, an additional LM component was added, trained on 133M word of LDC
Broadcast News texts.  The N-gram LM configuration is modeled after that described in \cite{saonSRK16}, except that
maxent smoothing was used.
The RNN and LSTM LMs were trained on Switchboard and Fisher transcripts as in-domain data (20M words for gradient computation,
3M for validation).
To this we added 62M words of UW Web data as out-of-domain data,
for use in the two-phase training procedure described above.  
\subsection{RNN-LM and LSTM-LM performance}
Table \ref{tab:rnnlm-results} gives perplexity and word error performance for various recurrent neural net LM setups,
from simple to more complex.  The acoustic model used was the ResNet CNN.
Note that, unlike the results in Tables~\ref{tab:lstmlm-layers} and~\ref{tab:lstmlm-perfs},
the neural net LMs in Table~\ref{tab:rnnlm-results} are interpolated with the N-gram LM.
As can be seen, each of the measures described earlier adds incremental gains, which, however, add up to a
substantial improvement overall.
The total gain relative to a purely N-gram based system is a 20\% relative error reduction with RNN-LMs,
and 23\% with LSTM-LMs.
As shown later (see Table~\ref{tab:main}) the gains with different acoustic models are similar.
\subsection{System Combination}
	\label{sec:combination}
The LM rescoring is carried out separately for each acoustic model.
The rescored N-best lists from each subsystem are then aligned into a single confusion 
network \cite{sri-2000} using the SRILM {\em nbest-rover} tool.
However, the number of potential candidate systems is too large to allow an all-out combination,
both for practical reasons and due to overfitting issues.
Instead, we perform a greedy search, starting with the single best system, and 
successively adding additional systems,
to find a small set of systems that are maximally complementary.
The RT-02 Switchboard set was used for this search procedure.
We experimented with two search algorithms to find good subsets of systems.
We always start with the system giving the best individual accuracy on the development set.
In one approach, a greedy forward search then adds systems incrementally to the combination, giving each equal
weight.  If no improvement is found with any of the unused systems, we try adding each with successively
lower relative weights of 0.5, 0.2, and 0.1, and stop if none of these give an improvement.
A second variant of the search procedure that can give lower error (as measured on the devset)
estimates the best system weights for each incremental combination candidate.
The weight estimation is done using an expectation-maximization algorithm based on aligning the reference words 
to the confusion networks, and maximizing the weighted probability of the correct word at each alignment position.
To avoid overfitting, the weights for an $N$-way combination are smoothed hierarchically, i.e., interpolated with the
weights from the $(N-1)$-way system that preceded it.
This tends to give robust weights that are biased toward the early (i.e., better) subsystems.
The final system incorporated a variety of BLSTM models with roughly similar performance,
but differing in various metaparameters
(number of senones, use of spatial smoothing, and choice of pronunciation dictionaries).%
\footnote{We used two different dictionaries, one based on a standard phone set and another with
dedicated vowel and nasal phones used only in the pronunciations of filled pauses (``uh'', ``um'')
and backchannel acknowledgments (``uh-huh'', ``mhm''), similar to \cite{sri-2000}.}
To further limit the number of free parameters to be estimated in system combination, we performed system selection
in two stages.  First, we selected the four best BLSTM systems.
We then combined these with equal weights and treated them as a single subsystem in searching for
a larger combination including other acoustic models.
This yielded our best overall combined system, as reported in Section~\ref{sec:compare}.
\section{Microsoft Cognitive Toolkit (CNTK)}
\label{sec:cntk}
All neural networks in the final system were trained with the
Microsoft Cognitive Toolkit, or CNTK \cite{cntkai,CNTK}.
on a Linux-based multi-GPU server farm.
CNTK allows for flexible model definition, while at the same
time scaling very efficiently
across multiple GPUs {\em and} multiple servers.
The resulting fast experimental turnaround
using the full 2000-hour corpus was critical for our work.
\subsection{Flexible, Terse Model Definition}
In CNTK, a neural network (and the training criteria) are specified by
its formula, using a custom functional language (BrainScript), or Python.
A graph-based execution engine, which provides automatic differentiation, then trains the model's parameters through SGD.
Leveraging a stock library of common layer types, networks can be specified
very tersely.
Samples can be found in \cite{cntkai}.
\subsection{Multi-Server Training using 1-bit SGD}
Training the acoustic models in this paper on a single GPU would take many weeks or even months.
CNTK made training times feasible by parallelizing the SGD training with our {\em 1-bit SGD}
parallelization technique \cite{seide20141}. This data-parallel method distributes minibatches over multiple worker nodes, and then aggregates the sub-gradients.
While the necessary communication time would otherwise be prohibitive,
the 1-bit SGD method eliminates the bottleneck by two techniques: {\em 1-bit quantization of gradients} and {\em automatic minibatch-size scaling}.
In \cite{seide20141}, we showed that gradient values can be quantized to just a single bit, if one carries over the quantization error from one minibatch to the next. Each time a sub-gradient is quantized, the quantization error is computed and remembered, and then added to the next minibatch's sub-gradient. This reduces the required bandwidth 32-fold with minimal loss in accuracy.
Secondly, automatic minibatch-size scaling progressively decreases the frequency of model updates. At regular intervals (e.g. every 72h of training data), the trainer tries larger minibatch sizes on a small subset of data and picks the largest that maintains training loss.
These two techniques allow for excellent multi-GPU/server scalability,
and reduced the acoustic-model training times on 2000h from months to between 1 and 3 weeks,
making this work feasible.
\subsection{Computational performance}
\begin{table*}
	\centering
	\caption{Runtimes as factor of speech duration for various aspects of acoustic modeling and decoding,
		for different types of acoustic model}
	\label{tab:runtimes}
	\begin{tabular}{|l|c|c|c|c|c|}
	\hline
	Processing step		& Hardware	& DNN		& ResNet-CNN	& BLSTM		& LACE	\\
	\hline \hline
	AM training		& GPU		& 0.012		& 0.60		& 0.022		& 0.23	\\
	\hline
	AM evaluation		& GPU		& 0.0064	& 0.15		& 0.0081	& 0.081	\\
	\hline
	AM evaluation		& CPU		& 0.052		& 11.7		& n/a		& 8.47	\\
	\hline
	Decoding		& GPU		& 1.04		& 1.19		& 1.40		& 1.38	\\
	\hline
	\end{tabular}
\end{table*}
Table~\ref{tab:runtimes} compares the runtimes, as multiples of speech duration, of various processing steps
associated with the different acoustic model architectures
(figures for DNNs are given only as a reference point, since they are not used in our system).
Acoustic model (AM) training comprises the forward and backward dynamic programming passes, as well as parameter updates.
AM evaluation refers to the forward computation only.
Decoding includes AM evaluation along with hypothesis search (only the former makes use of the GPU).
Runtimes were measured on a 12-core Intel Xeon E5-2620v3 CPU clocked at 2.4GHz, with an Nvidia Titan X GPU.
We observe that the GPU gives a 10 to 100-fold speedup for AM evaluation over the CPU implementation.
AM evaluation is thus reduced to a small faction of overall decoding time, making near-realtime operation possible.
\section{Experiments and Results}
\label{sec:results}
\subsection{Speech corpora}
We train with the commonly used English CTS (Switchboard and Fisher) corpora.
Evaluation is carried out on the NIST 2000 CTS test set, which comprises both Switchboard (SWB) and CallHome (CH)
subsets.
The waveforms were segmented according to the NIST partitioned evaluation map (PEM) file,
with 150ms of dithered silence padding added in the case of the CallHome conversations.%
\footnote{Using the {\tt sox} tool options {\tt pad 0.15 0.15 dither -p 14}.}
The Switchboard-1 portion of the NIST 2002 CTS test set was used for tuning and development.
The acoustic training data is comprised by LDC corpora 97S62, 2004S13, 2005S13,
2004S11 and 2004S09; see \cite{chen2006advances} for a full description.
\begin{table*}[t]
    \centering
    \caption{Word error rates (\%) on the NIST 2000 CTS test set with different acoustic models.
	Unless otherwise noted, models are trained on the full 2000 hours of data and have 9k senones.}
\vspace*{0.1in}
    \label{tab:main}
    \begin{tabular}{|l|l|l|l|l|l|l|}
    \hline
                Model  & \multicolumn{2}{|c|} {N-gram LM} & \multicolumn{2}{c|} {RNN-LM} & \multicolumn{2}{c|} {LSTM-LM}  \\ \hline \hline
               & CH & SWB                 & CH & SWB    & CH & SWB      \\ \hline
        ResNet, 300h training   & 19.2 & 10.0   & 17.7 & 8.2    & 17.0 & 7.7    \\      
        \hline
        ResNet                  & 14.8 & 8.6    & 13.2 & 6.9    & 12.5 & 6.6    \\      
        \hline
        ResNet, GMM alignments  & 15.3 & 8.8    & 13.7 & 7.3    & 12.8 & 6.9    \\      
        \hline
        VGG                     & 15.7 & 9.1    & 14.1 & 7.6    & 13.2 & 7.1    \\      
        \hline
        VGG + ResNet            & 14.5 & 8.4    & 13.0 & 6.9    & 12.2 & 6.4    \\      
        \hline
        
	% use LACE trained with 1/3 acweight because it works better in combo
        LACE                    & 15.0 & 8.4    & 13.5 & 7.2    & 13.0 & 6.7    \\      
        \hline
        
	% use this one instead because it's picked for the combo
        BLSTM                   & 16.5 & 9.0    & 15.2 & 7.5    & 14.4 & 7.0    \\      
        \hline
        BLSTM, spatial smoothing
				& 15.4 & 8.6    & 13.7 & 7.4    & 13.0 & 7.0    \\      
        \hline
        BLSTM, spatial smoothing, 27k senones
			   	& 15.3 & 8.3    & 13.8 & 7.0   & 13.2 & 6.8    \\      
       \hline
        BLSTM, spatial smoothing, 27k senones, alternate dictionary
                                & 14.9 & 8.3    & 13.7 & 7.0    & 13.0 & 6.7    \\      
        \hline \hline
	BLSTM system combination
				& 13.2	& 7.3	& 12.1	& 6.4 	& 11.6 & 6.0	\\	% expt235, expt237, expt232
        \hline \hline
        Full system combination & 13.0	& 7.3   & 11.7 & 6.1    & {\bf 11.0} & {\bf 5.8}    \\ 
        \hline 
    \end{tabular}
\end{table*}
\begin{table}
    \centering
    \caption{Comparative error rates from the literature and human error as measured in this work}
    \vspace*{0,1in}
    \label{tab:comps}
    \begin{tabular}{|l|l|l|l|l|}
    \hline
        \multirow{2}{*}{Model}  & \multicolumn{2}{c|} {N-gram LM} & \multicolumn{2}{c|} {Neural net LM}  \\ \cline{2-5}
                                & CH & SWB      & CH & SWB      \\
        \hline \hline
        Povey et al. \cite{povey2016purely} LSTM        & 15.3  & 8.5  & -  & - \\
        \hline
        Saon et al. \cite{saonSRK16} LSTM        & 15.1  & 9.0  & -  & -    \\
        \hline
        Saon et al. \cite{saonSRK16} system      & 13.7  & 7.6  & 12.2 & 6.6    \\
        \hline
	2016 Microsoft system	  & 13.3 & 7.4	& 11.0 & 5.8 \\
        \hline \hline
        Human transcription       &    &  	& 11.3 & 5.9 \\
	\hline
    \end{tabular}
\end{table}
\subsection{Acoustic Model Details}
Forty-dimensional log-filterbank features were extracted every 10 milliseconds,
using a 25-millisecond analysis window. The CNN models used window sizes
as indicated in Table~\ref{tab:CNNs}, and the LSTMs processed one frame
of input at a time. The bulk of our models
use three state left-to-right triphone models 
with 9000 tied states. Additionally, we have trained several models with
27k tied states. The phonetic 
inventory includes special models for 
noise, vocalized-noise, laughter and silence. We use a 30k-vocabulary
derived from the most common words in the Switchboard and Fisher corpora.
The decoder uses a statically compiled unigram graph, and dynamically
applies the language model score. The unigram graph has about 300k states and
500k arcs. 
Table \ref{tab:lfmmi} shows the result of i-vector adaptation and LFMMI training
on several
of our early systems. We achieve a 5--8\% relative improvement from i-vectors, including on
CNN systems. 
The last row of Table \ref{tab:lfmmi} shows the effect of LFMMI training on the 
different models. We see a consistent 7--10\% further 
relative reduction in error
rate for all models. Considering the great increase in procedural simplicity of
LFMMI over the previous practice of writing lattices and post-processing them,
we consider LFMMI to be a significant advance in technology.
\subsection{Overall Results and Discussion}
\label{sec:compare}
The performance of all our component models 
is shown in Table~\ref{tab:main}, along with the BLSTM combination and full system combination results.
(Recall that the four best BLSTM systems are combined with equal weights first,
as described in Section~\ref{sec:combination}.)
Key benchmarks from the literature, our own best results, and the measured human error rates are compared in
Table~\ref{tab:comps}.%
\footnote{When comparing the last row in Table~\ref{tab:lfmmi} with the ``N-gram LM'' results in Table~\ref{tab:main},
note that the former results were obtained with the pruned N-gram LM used in the decoder and
fixed score weights (during lattice generation),
whereas the latter results are from rescoring with the unpruned N-gram LM (during N-best generation),
using optimized score weighting.  Accordingly, the rescoring results are generally somewhat better.}
All models listed in Table~\ref{tab:main} are selected for the combined systems
for one or more of the three rescoring LMs.
The only exception is the VGG+ResNet system, which combines acoustic senone posteriors from the VGG and ResNet networks.
While this yields our single best acoustic model, only the individual VGG and ResNet models are used in the
overall system combination.
We also observe that the four model variants chosen for the combined BLSTM subsystem differ incrementally by
one hyperparameter (smoothing, number of senones, dictionary), and that the BLSTMs alone achieve an error
that is within 3\% relative of the full system combination.
This validates the rationale that choosing different hyperparameters is an effective way to obtain complementary
systems for combination purposes.
We also assessed the lower bound of performance for our lattice/N-best rescoring paradigm.
The 500-best lists from the lattices generated with the ResNet CNN system
had an oracle (lowest achievable) WER of 2.7\% on the Switchboard portion of the NIST 2000 evaluation set,
and an oracle WER of 4.9\% on the CallHome portion.
The oracle error of the combined system is even lower (though harder to quantify) since (1) N-best output from all 
systems are combined and (2) confusion network construction generates new possible hypotheses not contained
in the original N-best lists.
With oracle error rates less than half the currently achieved actual error rates,
we conclude that search errors are not a major limiting factor to even better accuracy.
\section{Error Analysis}
\label{sec:analysis}
In this section, we compare the errors made by our artificial recognizer
with those made by human transcribers.
We find that the machine errors
are substantially the same as human ones, with one large exception: confusions
between {\it backchannel} words and {\it hesitations}. The distinction is that
backchannel words like ``uh-huh'' are an acknowledgment of the speaker, also signaling 
that the speaker should keep talking,
while hesitations like ``uh'' are used to indicate that 
the current speaker has more to say and wants to keep his or her turn.%
\footnote{The NIST scoring protocol treats hesitation words as optional, and we therefore 
delete them from our recognizer output prior to scoring.  This explains why we see many substitutions of 
backchannels for hesitations, but not vice-versa.}
As turn-management devices, these two classes of words therefore have exactly opposite functions.
\begin{table*}[t]
    \centering
\caption{Most common substitutions for ASR system and humans. The number of
times each error occurs is followed by the word in the reference, and what
appears in the hypothesis instead.}
\vspace*{0.1in}
\label{tab:subs}
        \small
    \begin{tabular}{|l|l||l|l|}
    \hline
    \multicolumn{2}{|c||}{{\bf CH}} & \multicolumn{2}{c|}{{\bf SWB}} \\ \hline
                    {\bf ASR}   & {\bf Human}   & {\bf ASR}    & {\bf Human}   \\ \hline \hline
45:  (\%hesitation) / \%bcack & 12:  a / the & 29:  (\%hesitation) / \%bcack & 12:  (\%hesitation) / hmm \\ \hline
12:  was / is & 10:  (\%hesitation) / a &  9:  (\%hesitation) / oh & 10:  (\%hesitation) / oh \\ \hline
9:  (\%hesitation) / a & 10:  was / is &  9:  was / is &  9:  was / is \\ \hline
8:  (\%hesitation) / oh & 7:  (\%hesitation) / hmm &  8:  and / in &  8:  (\%hesitation) / a \\ \hline
8:  a / the & 7:  bentsy / bensi &  6:  (\%hesitation) / i &  5:  in / and \\ \hline
7:  and / in & 7:  is / was &  6:  in / and &  4:  (\%hesitation) / \%bcack \\ \hline
7:  it / that & 6:  could / can &  5:  (\%hesitation) / a &  4:  and / in \\ \hline
6:  in / and & 6:  well / oh &  5:  (\%hesitation) / yeah &  4:  is / was \\ \hline
5:  a / to & 5:  (\%hesitation) / \%bcack &  5:  a / the &  4:  that / it \\ \hline
5:  aw / oh & 5:  (\%hesitation) / oh &  5:  jeez / jeeze &  4:  the / a \\ \hline
        \end{tabular}
\end{table*}
\begin{table}[t]
    \centering
\caption{Most common deletions for ASR system and humans.}
\vspace*{0.1in}
\label{tab:dels}
        \small
    \begin{tabular}{|l|l||l|l|}
    \hline
    \multicolumn{2}{|c||}{{\bf CH}} & \multicolumn{2}{c|}{{\bf SWB}} \\ \hline 
                    {\bf ASR}   & {\bf Human}   & {\bf ASR}    & {\bf Human}   \\ \hline \hline
 44:  i  &     73:  i &    31:  it &    34:  i \\ \hline
      33:  it &       59:  and &       26:  i &       30:  and \\ \hline
      29:  a &       48:  it &       19:  a &       29:  it \\ \hline
      29:  and &       47:  is &       17:  that &       22:  a \\ \hline
      25:  is &       45:  the &       15:  you &       22:  that \\ \hline
      19:  he &       41:  \%bcack &       13:  and &       22:  you \\ \hline
      18:  are &       37:  a &       12:  have &       17:  the \\ \hline
      17:  oh &       33:  you &       12:  oh &       17:  to \\ \hline
      17:  that &       31:  oh &       11:  are &       15:  oh \\ \hline
     17:  the &      30:  that &      11:  is &      15:  yeah \\ \hline
        \end{tabular}
\end{table}
\begin{table}[t]
    \centering
\caption{Most common insertions for ASR system and humans.}
\vspace*{0.1in}
\label{tab:ins}
        \small
    \begin{tabular}{|l|l||l|l|}
    \hline
    \multicolumn{2}{|c||}{{\bf CH}} & \multicolumn{2}{c|}{{\bf SWB}} \\ \hline 
                    {\bf ASR}   & {\bf Human}   & {\bf ASR}   & {\bf Human}  \\ \hline \hline
15:  a &    10:  i &    19:  i &    12:  i \\ \hline
      15:  is &        9:  and &        9:  and &       11:  and \\ \hline
      11:  i &        8:  a &        7:  of &        9:  you \\ \hline
      11:  the &        8:  that &        6:  do &        8:  is \\ \hline
      11:  you &        8:  the &        6:  is &        6:  they \\ \hline
       9:  it &        7:  have &        5:  but &        5:  do \\ \hline
       7:  oh &        5:  you &        5:  yeah &        5:  have \\ \hline
       6:  and &        4:  are &        4:  air &        5:  it \\ \hline
       6:  in &        4:  is &        4:  in &        5:  yeah \\ \hline
      6:  know &       4:  they &       4:  you &     4:  a \\ \hline
        \end{tabular}
\end{table}
Table \ref{tab:subs} shows the ten most common substitutions for both
humans and the artificial system. Tables \ref{tab:dels} and \ref{tab:ins}
do the same for deletions and insertions. Focusing on the substitutions,
we see that by far the most common error in the ASR system is the confusion
of a hesitation in the reference for a backchannel in the hypothesis.
People do not seem to have this problem.  We speculate that this is due
to the nature of the Fisher training corpus, where the ``quick transcription''
guidelines were predominately used \cite{glenn2010transcription}. 
We find that there is inconsistent treatment of backchannel and hesitation
in the resulting data; the relatively poor performance of the 
automatic system here might simply 
be due to confusions in the training data annotations.
For perspective, there are over twenty-one thousand words in each 
test set. Thus the errors due to hesitation/backchannel substitutions account
for an error rate of only about 0.2\% absolute.
 
The most frequent
substitution 
for people on the Switchboard corpus was mistaking  a hesitation in the
reference for the word ``hmm.'' 
The scoring guidelines treat ``hmm''
as a word distinct from backchannels and hesitations, so this is not
a scoring mistake. Examination of the contexts in which the error is made 
show that it is most often intended to acknowledge the other speaker,
i.e. as a backchannel. For both people and our automated system, the
insertion and deletion patterns are similar: short function words are
by far the most frequent errors. 
In particular, the single most common error made by the transcribers
was to omit the word ``I.''
While we believe further improvement in 
function and content words is possible, the significance of the
remaining backchannel/hesitation confusions is unclear.
Table \ref{tab:dtl} shows the overall error rates broken down by 
substitutions, insertions and deletions. We see that the human transcribers
have a somewhat lower substitution rate, and a higher deletion rate. The
relatively higher deletion rate might reflect a human bias to avoid 
outputting uncertain information, or the productivity demands on a 
professional transcriber. In all cases, the number of insertions is 
relatively small.
 
\begin{table}[t]
    \centering
\caption{Overall substitution, deletion and insertion rates.}
\vspace*{0.1in}
\label{tab:dtl}
        \small
    \begin{tabular}{|l|l|l||l|l|}
    \hline
    & \multicolumn{2}{|c||}{{\bf CH}} & \multicolumn{2}{c|}{{\bf SWB}} \\ \hline 
            &         {\bf ASR}   & {\bf Human}   & {\bf ASR}   & {\bf Human}  \\ \hline \hline
sub & 6.5 & 4.1 & 3.3 &  2.6     \\ \hline
del & 3.3 & 6.5 & 1.8 &  2.7     \\ \hline
ins & 1.4 & 0.7 & 0.7 &  0.7     \\ \hline
all & 11.1 & 11.3 & 5.9 & 5.9    \\ \hline
        \end{tabular}
\end{table}
\section{Relation to Prior Work}
\label{sec:prior}
Compared to earlier applications of CNNs to speech recognition
\cite{sainath2013deep,abdel2012applying}, our networks are 
much deeper, and use linear bypass connections across convolutional
layers. They are similar in spirit to those studied more recently by
\cite{sercu2016very,saon2015ibm,saonSRK16,bi2015very,qian2016very}. We improve on these 
architectures with the LACE model \cite{yu2016deep}, which iteratively
expands the effective window size, layer-by-layer, and adds an
attention mask to differentially weight distant context.
Our spatial regularization technique is similar in spirit to stimulated
deep neural networks~\cite{wu2016stimulated}. Whereas stimulated networks
use a supervision signal to encourage locality of activations in the model,
our technique is automatic.
Our use of 
lattice-free MMI is distinctive, and extends previous work
\cite{chen2006advances,povey2016purely} by proposing the use of a mixed
triphone/phoneme history in the language model.
On the language modeling side, we achieve a performance boost by combining multiple LSTM-LMs 
in both forward and backward directions, and by using a two-phase training regimen to get
best results from out-of-domain data.
For our best CNN system, LSTM-LM rescoring yields a relative word error reduction of 23\%,
and a 20\% relative gain for the combined recognition system, considerably larger than previously 
reported for conversational speech recognition \cite{medennikov2016improving}.
\section{Conclusions}
\label{sec:concl}
We have measured the human error rate on NIST's 2000 conversational telephone speech recognition
task. We find that there is a great deal of variability between the 
Switchboard and CallHome subsets, with 5.8\% and 11.0\% error rates 
respectively. For the first time, we report automatic recognition performance on par with
human performance on this task. 
Our system's performance can be attributed to the systematic use of LSTMs for
both acoustic and language modeling, as well as CNNs in the acoustic
model, and extensive combination of complementary system for both acoustic and language modeling.
\section*{Acknowledgments}
We thank Arul Menezes for access to the Microsoft transcription pipeline;
Chris Basoglu, Amit Agarwal and Marko Radmilac for their invaluable 
assistance with CNTK; Jinyu Li and Partha Parthasarathy for many helpful
conversations.
We also thank X. Chen from Cambridge University for valuable assistance with the CUED-RNNLM toolkit,
and the International Computer Science Institute for compute and data resources.
\bibliographystyle{ieee-shortnames}
\bibliography{strings,refs}
\end{document}